\pdfoutput=1

\documentclass[11pt]{article}
\usepackage[table,xcdraw]{xcolor}
\usepackage[preprint]{acl}
\usepackage{fvextra}
\usepackage{pifont}
\usepackage{times}
\usepackage{latexsym}
\usepackage{listings}
\usepackage{xcolor}
\usepackage[T1]{fontenc}

\usepackage[utf8]{inputenc}

\usepackage{microtype}

\usepackage{inconsolata}

\usepackage{graphicx}
\usepackage{booktabs}

\newcommand{\ensuretext}[1]{#1}
\newcommand{\marker}[2]{\ensuremath{^{\textsc{#1}}_{\textsc{#2}}}}

\newcommand{\draftcomment}[3]{\ensuretext{\textcolor{#3}{[#1 #2]}}}

\newcommand{\kexin}[1]{\draftcomment{\marker{K}{P}}{#1}{purple}}
\usepackage{xspace}
\newcommand{\name}{PoPilot\xspace}

\usepackage{listings}

\title{Building A Proof-Oriented Programmer That Is 64\% Better Than GPT-4o Under Data Scarcity} 

\author{
 \textbf{Dylan Zhang\textsuperscript{1,\ddag}},
 \textbf{Justin Wang\textsuperscript{2}},
 \textbf{Tianran Sun\textsuperscript{3}}
\\
 \textsuperscript{1}University of Illinois Urbana-Champaign, \\
 \textsuperscript{2}University of Chicago,\\
 \textsuperscript{3}Shanghai Jiaotong University,\\
 \small{\textsuperscript{\ddag}Project Lead}\\
 \small{
   \textbf{Correspondence:} \href{mailto:shizhuo2@illinois.edu}{shizhuo2@illinois.edu}
 }
}

\begin{document}
\maketitle
\begin{abstract}
Existing LMs struggle with proof-oriented programming due to data scarcity, which manifest in two key ways: (1) a lack of sufficient corpora for proof-oriented programming languages such as F*, and (2) the absence of large-scale, project-level proof-oriented implementations that can teach the model the intricate reasoning process when performing proof-oriented programming. We present the first  on synthetic data augmentation for project level proof oriented programming for both generation and repair. Our method addresses data scarcity by synthesizing basic proof-oriented programming problems for proficiency in that language; incorporating diverse coding data for reasoning capability elicitation and creating new proofs and repair data within existing repositories. This approach enables language models to both synthesize and repair proofs for function- and repository-level code. We show that our fine-tuned 14B parameter model, \name, can exceed the performance of the models that 
outperforms GPT-4o in project-level proof-oriented programming by 64\% relative margin, and can improve GPT-4o's performance by 54\% by repairing its outputs over GPT-4o's self-repair.

\end{abstract}
\section{Introduction}
In an era where software vulnerabilities can risk tremendous damages\footnote{On July 19, 2024, an abnormal update distributed by the cybersecurity company CrowdStrike caused issues for a large number of its global customers’ Windows operating systems, resulting in blue screen errors.} and compromise national security \footnote{U.S. Cybersecurity and Infrastructure Security Agency has urged the future software development to ensure memory safety.}, ensuring the correctness and safety has become an urgent priority. Proof-oriented programming integrates formal verification into software development, enabling mathematically rigorous correctness guarantees.
 Proof-oriented programming languages such as F*\cite{swamy2011secure} and Dafny\cite{dafny} support this paradigm by providing expressive type systems and precise specification mechanisms. These capabilities enable static verification of program correctness without the need for extensive test suites or runtime execution. These languages like F* allow developers to write programs alongside formal proofs, providing strong guarantees about functional correctness and security properties, and famous real-world systems including Firefox, the Linux kernel, Tezos blockchain, and Azure Cloud have employed formally verified components.
 However, despite decades of research, the adoption of proof-oriented programming remains limited due to the high cost of proof construction and the steep learning curve of formal methods.

On the other hand, the rapid advancements in large language models (LLMs) have transformed many areas of software development\cite{austin2021program, chen2023teaching, nijkamp2022codegen, xia2022less, xia2023automated, jin2023inferfix, jimenez2024swebenchlanguagemodelsresolve}. 
Yet, applying LLMs to proof-oriented programming presents fundamental challenges~\cite{chakraborty2024towards}. Proof-oriented programming is highly distinct from conventional coding paradigms, demanding complex formal reasoning of program semantics over long contexts, which current models struggle with~\cite{loughridge2024dafnybench}. A major bottleneck is extreme data scarcity, which manifests in two key ways: (1) a lack of diverse, high-quality corpora in proof-oriented programming languages such as F* to teach the model the syntax and semantics of that language (F* constitutes of only 0.002\% in Stack-v2 ~\cite{lozhkov2024starcoder}), and (2) an absence of large-scale project-level verification data, which involves highly complex and context-dependent formal reasoning. As a result, even state-of-the-art LMs fail to generalize effectively to proof construction and verification tasks~\cite{loughridge2024dafnybench,fstar2024}.

In this work, we introduce a data-centric post-training recipe designed to bridge the gap between general-purpose coding LLMs and repository-level proof-oriented programming in F*. We systematically address data scarcity and adaptation challenges through three key strategies:
\begin{enumerate}
    \item \textbf{Enhancing General Programming Capabilities with Diverse Code Data:} inspired by existing works showing the benefit of diversification~\cite{zhang2024textbf,dong2023abilities,chen2024diversity}, we train the model beyond the immediate focus of formal verification in F* but over diverse programming tasks to enable its code-reasoning and instruction-following capabilities. 
    \item  \textbf{Learning Basic Proof-Oriented Programming via Synthetic Tasks:}
We synthesize function-level basic programming and property-proving problems in F*, allowing LLMs to learn fundamental verification patterns in a controlled setting without introducing complex inter-dependencies.
\item  \textbf{Synthetic Augmentation for Proof Synthesis and Repair:}
Beyond training basic property-proving problems and existing repositories, we curate novel synthetic repository-level problem solving and proof repair data to teach LLMs how to complete and correct project-level proofs with more complex and longer contextual dependencies.
\end{enumerate}
Following the recipe above, we transform LLMs into specialized verification assistants capable of both synthesis and repair for \textbf{P}roof-\textbf{o}riented \textbf{P}rogramming, which we call \name. \name is the first project-level formal verification specialist LLM trained on synthetic instruction-tuning data. 
Notably, strategies \textbf{2} and \textbf{3} require generating problems using existing language models that has limited knowledge and low accuracy on F*. 
However, we could leverage F* solver to obtain data with correctness guarantee for generation tasks, and retrieve error messages to craft repair datasets\footnote{The correctness can be determined by running the solver without test cases as in conventional programming languages.}
Our experiments demonstrate that \name demonstrates strong capacities to perform proof-oriented programming on a project level, leading to a remarkable margin of 64\% over GPT-4o and can boost GPT-4o's performance to 54\% by repairing a randomly chosen failed attempt.

\section{Background}
\begin{figure*}[h]
    \centering
    \includegraphics[width=.8\linewidth]{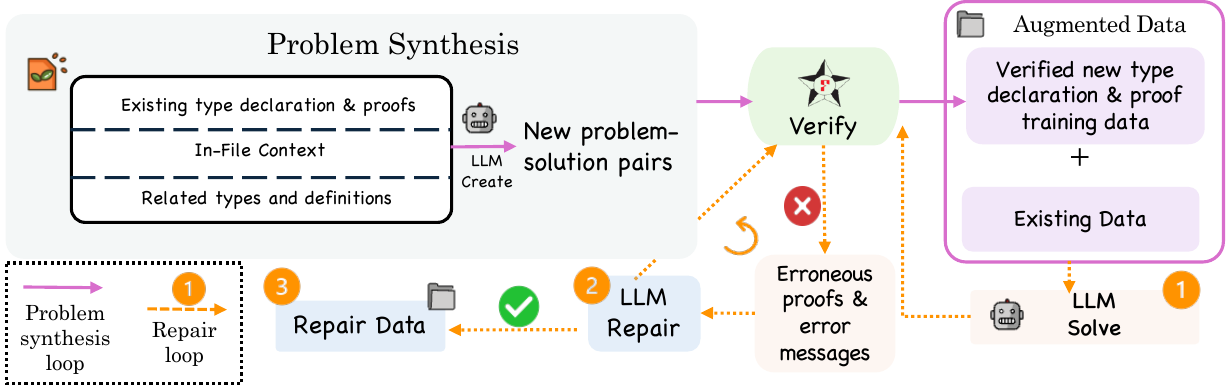}
    \caption{Illustration of repository-level data generation pipeline.}
    \label{fig:pipline}
\end{figure*}
Formal verification is a process of examining the correctness of the operation of software programs by mathematical proof~\cite{digitalsystem}. F* is an SMT-solver based proof-oriented language that enables convenient verification through execution by F* compiler. However, F*, like many other languages used for formal proofs (verified Rust, Rocq(Coq), Lean etc.), are comparatively low resource.  The popular open coding data corpus Stack-v2~\cite{lozhkov2024starcoder} containing over 3B files in 600+ programming and markup languages,  F* has only 29.6k entries (less than 0.002\%), much fewer than common programming languages like Python (80.6M entries, 2.95\%), Java (223M entries, 8.17\%). This scarcity implies both the lack of knowledge of the off-the-shelf pre-trained checkpoints and the insufficiency of existing resources to further train the model. 

Additionally, F* is a dependently-typed language, where type definitions depend on values. This allows for more precise specification of program properties and invariants but also introduces complexity due to the need for intricate computations to determine type equality and detailed type reasoning \cite{chakraborty2024towards}. Programmers also often need to go back and forth while writing F* code. Therefore, enhancing the model's ability to repair code is crucial for both iterative improvement of automatic proof synthesis and better assistance to human programmers.


In reality, the challenge of proof-oriented programming is further exacerbated by the cross-repository dependencies of the code. Proof-oriented programming often spans multiple repositories, especially in the large-scale formal verification of software systems, where the project contains different components relying on verifying properties from other repositories. This introduces huge challenges in dependency and environment management, as the proofs account for resolving inter-repository specification and module openings beyond the single repository \cite{chakraborty2024towards}. This property, together with the type-dependent nature of the programming language, makes the task difficult to learn for model and even human expertise. 
\section{Function-Level Dataset Collection}
In this section, we describe how we synthesize diverse function-level programming tasks from existing open-source code snippets. We focus on three types of tasks: natural language to code tasks, proof-oriented code completion tasks, and code repair tasks. To evaluate the generated data, we rely on F*’s feature to automatically verify code correctness via execution. In addition, the dataset is diversified by incorporating instruction-tuning data from other languages.




\subsection{Data Synthesis}
In this section, we describe our method for synthesizing high-quality F* training data. To generate diverse programming tasks, we construct instruction data from existing open-source code snippets inspired by OSS-Instruct~\cite{wei2024magicoder}. We select code models that outperform non-code models in terms of accuracy and quality for F* code generation and prompt them to create high-quality tasks from a variety of code examples.

\subsubsection{Task generation}
We included three types of tasks in our dataset: natural language based tasks, proof-oriented code completion tasks and code repair tasks. Below, we describe each task in detail.

\paragraph{Natural Langugage To Code Tasks}

We use code examples from the F* source code repository and GitHub OCaml ~\cite{ocamlgithub} as our seed corpus.The code snippets extracted from these source codes are diverse, correct and practical, inspiring LLM to generate high-quality and varied instructions and responses. The response code generated by the model is also required to be self-contained to facilitate the verification of code correctness and the generation of other types of tasks. 

\paragraph{Proof-oriented code completion tasks}
In proof-oriented code completion problems, we choose verified and self-contained code snippets from the responses models generated in all tasks. We then ask the model to generate code completion problems that require proving a specific property or function related to the given snippet in order to obtain problems and corresponding responses related to proofs. This part of data, with greater complexity, enhances the model’s ability to write proofs in F* and perform reasoning. Verified response codes can also be reused as snippets for future code completion tasks.
\begin{figure}[htbp]
    \centering
    \includegraphics[width=\linewidth]{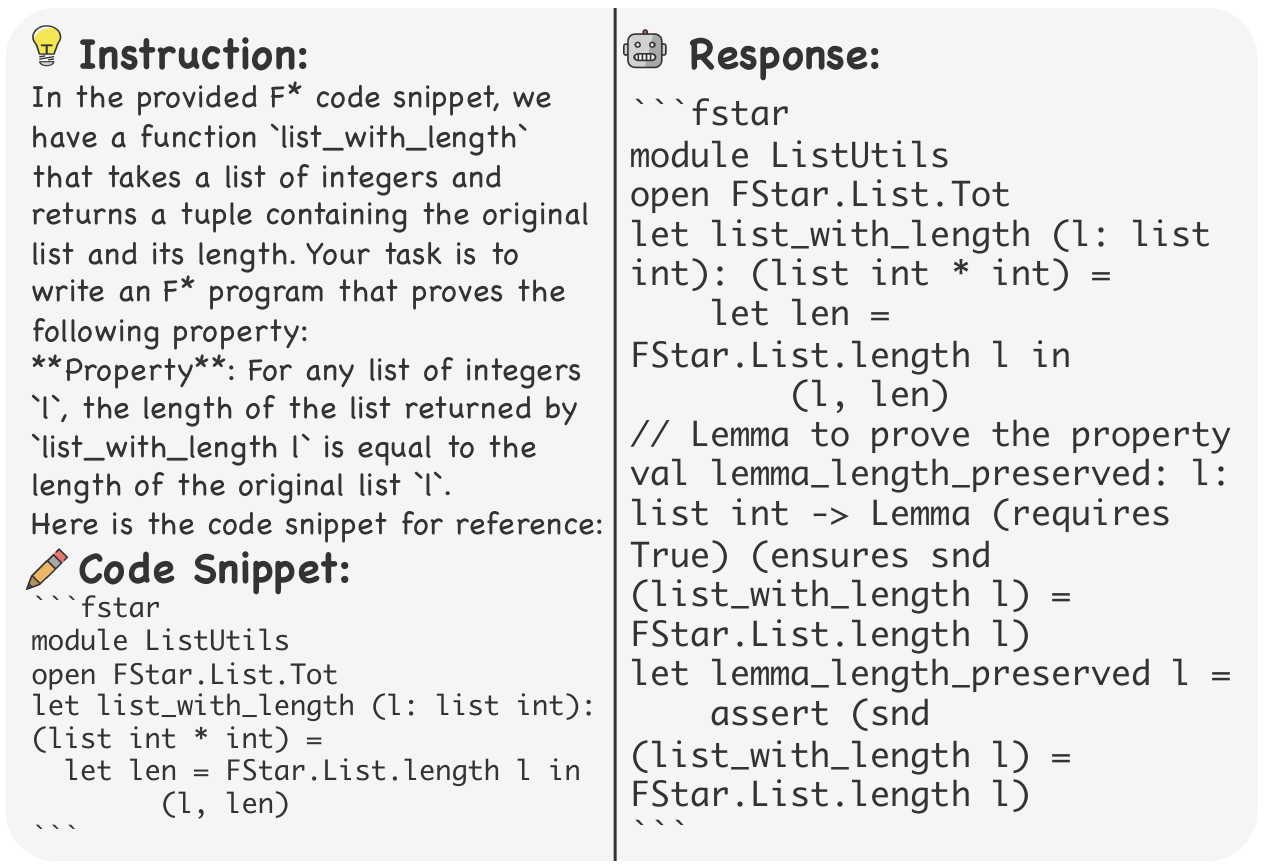}
    \caption{Function-level Proof-oriented programming example.}
    \label{fig:code2codeillu}
\end{figure}
\paragraph{Code repair tasks.}
It is difficult to generate correct code on the first try for many programming tasks, even for human. Instead of discarding the flawed code, they typically analyze the results and modify it to fix any errors~\cite{chen2023teaching}. Therefore, It is crucial for LLMs to have the capability to repair code as well.  Given the complexity of writing F* programs, which often requires iterative refinement, models trained on synthetic datasets should be capable of repairing erroneous code effectively. Code repair pairs in our dataset are generated by prompting the model to fix the given incorrect code response, supplying both the erroneous code and the execution log obtained after verification.

The verified instruction-response pairs are added to our dataset.

\subsubsection{Execution-based data evaluation and dataset filtering}\label{Execution-based data evaluation and dataset filtering}
As an SMT-guided language, F* can directly and accurately receive correctness feedback from execution, enabling the identification of discrepancies between code behavior and formal specifications. Therefore, no extra effort such as test cases and LLM judges is required for verifying F* data. Specifically, to verify the response codes, we put the code snippet within the F* environment and execute them, sparing little effort. Data that successfully compile and run are retained, while incorrect ones and their associated error messages are stored for inclusion in the repair dataset. This verifiability is a favorable property of F* that we could leverage to obtain data quality signals for free, whereas in general domains, the instruction tuning data is largely unverifiable or relies on heuristics~\cite{wei2024selfcodealign}. 

\subsection{Diversification} \label{Diversification}
Diverse instructions can better enhance an LLM’s ability to generalize to new tasks \cite{wei2021finetuned}, as well as improve its comprehension and adherence to instructions ~\cite{chen2024diversity,zhang2024textbf,dong2023abilities}. This is particularly crucial in our case since the F* community is smaller than common programming languages with scarcer source codes and more sparse documentation. Therefore, we integrate diverse instruction-tuning data pairs in other languages besides synthetic F* data to supplement our dataset and enhance model's capability. This general approach is applicable beyond F*, as it can be extended to other programming languages with limited available data, helping to improve model performance in low-resource scenarios.


\section{Project-Level Dataset Synthesis}

In this section, we describe how we synthesize more project-level proof generation problems from existing repositories. Project-level verification involves generating or repairing proofs for definitions embedded within complete verification repositories, where correctness depends not only on local function logic but also on broader module-level context—including previously established lemmas, type invariants, module imports, and shared assumptions. Our goal is to generate new problem-solution pairs based on existing programming contexts where the \textbf{problem} consists of a type declaration, and the \textbf{solution} is the correct F* definition (a proof) that satisfies the given type declaration (See examples in \ref{sec:Problem-Solution Pair Examples}).

\subsection{Generating New Problems} \label{Generating New Problems}

We start the problem-solution pairs generation from a seed dataset, in which each instance has the following structure:

    \paragraph{Definition}An initial problem-solution pair of a definition in the context. 
    \paragraph{Context} Required context information from existing repositories, such as opened premises and pre-defined definitions, and selected premises which are likely to be used in the body of the definitions\cite{yang2023leandojo}.
    \paragraph{Examples} A set of semantically similar problem-solution pairs retrieved from the same context using the similarity between types in their embedding space\cite{chakraborty2024towards}.

We prompt the language model to create new definitions or prove new properties based on the context in the seed dataset. The detail is listed as follows:

\paragraph{Generation Prompt Curation:} We structure the prompt with relevant premises and pre-defined definitions, providing essential context for generating new definitions and proofs. (Prompt see \ref{sec:New Definition Prompt Example}). To guide the model in following F* conventions, we retrieve multiple example definitions with varied structures from the same context while ensuring diversity and discouraging direct copying.

\paragraph{Definition Generation:} For each unique context in the seed dataset, we apply the predefined prompt template and sample multiple candidate problem-solution pairs using two LLMs.

\paragraph{Data Filtering:} To maintain quality and prevent redundancy, we apply de-duplication by (1) computing sequence similarity and filtering out overly similar definitions, including those resembling reference examples, and (2) remove any generated definitions that overlap with the test set to prevent data leakage and ensure a fair evaluation. 

\begin{figure}
    \centering
    \includegraphics[width=.75\linewidth]{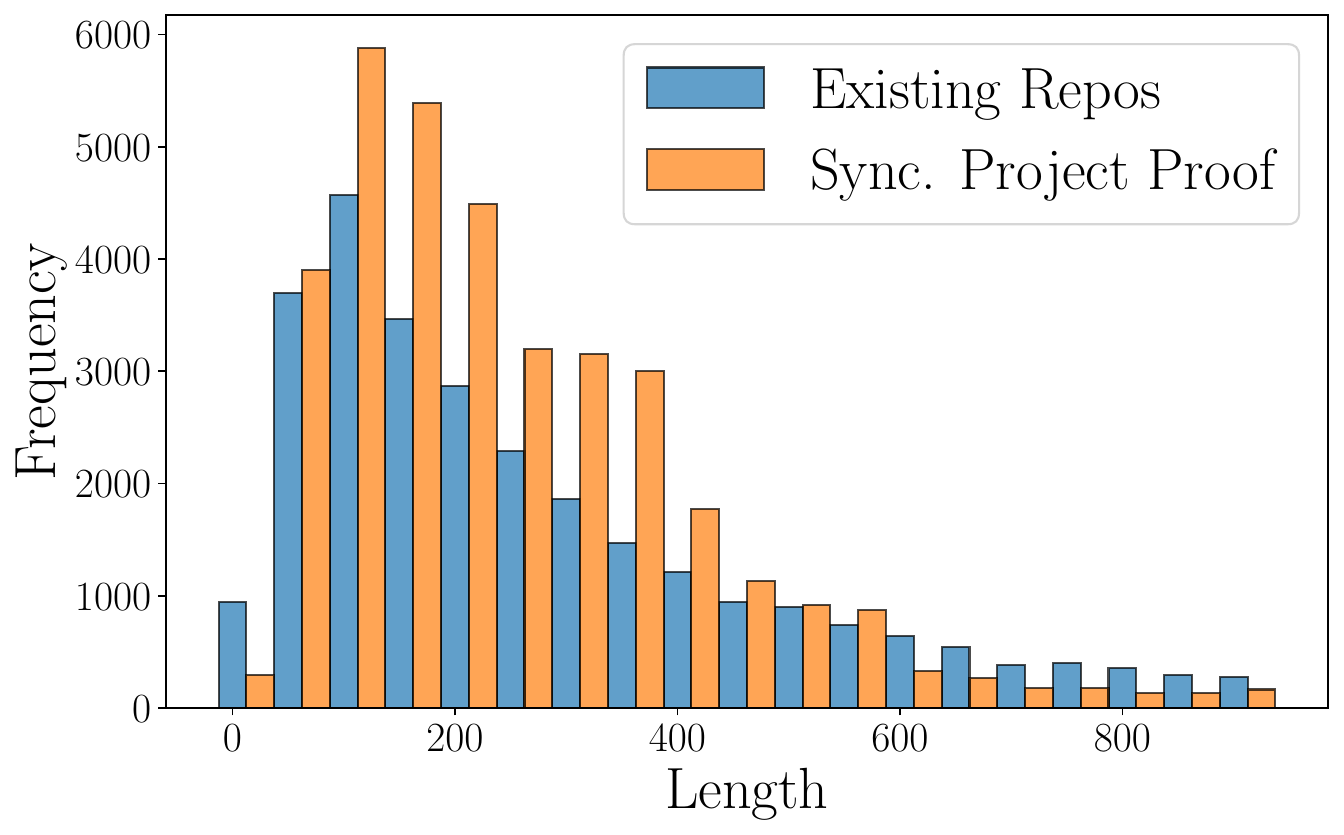}
    \caption{Length Comparison between Generated Definitions vs Existing Definitions.}
    \label{fig:length}
\end{figure}

These steps expand the dataset with diverse problem-solution pairs while maintaining real-world F* relevance and validation feasibility. Figure \ref{fig:length} shows that the problem-solution pair augmentation is effective for simpler definitions but struggles with longer and more complex ones, reflecting the long-tailed distribution observed in both real-world and synthesized datasets. This alignment suggests that our synthetic data captures the difficulty distribution in real F* development.

\subsection{Creating Repair Data}

In this section, we generate new problem-solution pairs for the proof repair task. The \textbf{problem} consists of a type declaration, an incorrect proof, and the corresponding error message from the F* compiler, while the \textbf{solution} is the corrected proof. To construct this dataset, we combine rule-based data synthesis with LLM-generated repair data.

\subsubsection{Synthetic Repair Data} \label{Synthetic Repair Data}
We generate a synthetic mutation dataset from the F* dataset by randomly modifying ground-truth solutions at the abstract syntax tree level. If a mutation causes type checking to fail, we treat it as a synthetic error and use it to train a model to recover the original solution. Mutations include omitting parts, replacing arguments with underscores, and modifying control structures (e.g., removing branches or let-definitions). These errors mimic those commonly made by human F* programmers, but we avoid mutating identifiers.
\subsubsection{Repair Data From The Model}\label{Repair Data From The Model}

Since Section \ref{Generating New Problems} expands the problem set, we now prompt the model to solve these problems within their given contexts, collecting incorrect proofs and their corresponding error messages. Correct answers for the repair task are retrieved either from the original correct proofs or by prompting LLMs to generate new valid proofs.

\paragraph{Repair Problems Generation:} We combine the seed dataset problems with the generated ones and prompt the LLM to solve them. Solutions are validated (\ref{Diversification}), and incorrect proofs with error messages are collected from the compiler as repair problems.

\paragraph{Obtaining Correct Repairs:} (1) We prompt LLMs to solve the repair problems directly, though zero-shot performance is often limited. (2) Alternatively, we reuse the original correct proofs from the definition generation task as repair solutions.

\paragraph{Data Filtering:} (1) Duplicate repair problems are removed by filtering identical incorrect proofs. (2) Each definition generation problem contributes at most three repair problems to prevent redundancy. (3) Correct answers appearing in the test set are removed to ensure fair evaluation.

We can see that the dominating model-generated errors is \textit{identifier not found} and \textit{syntax error}. While syntax errors reflect model's limited understanding of F* grammar, identifier not found errors indicate deeper semantic and type-related challenges that are characteristic of F* language.

\section{Function-Level Experiments}
We present our experiments on the function-level synthetic data mixtures in this section.
\subsection{Setup}
\paragraph{Dataset}
Our dataset primarily comprises synthetic F* data, along with selected data from datasets of other programming languages (CodeAlpaca, Evol-Instruct, Deepseek-Prover-V1, RunbugRun, stack-exchange-preferences). CodeAlpaca collects synthetic coding instructions from ChatGPT following self-instruct, including diverse programming problems and languages~\cite{codealpaca}. Evol-Instruct also contains high-quality synthetic natural language to code data~\cite{wei2024magicoder}. The Deepseek-Prover-V1~\cite{xin2024deepseek} dataset includes extensive Lean 4 proof data to enhance theorem-proving capabilities in LLMs. RunbugRun provides an executable dataset for automated program repair on commonly used programming languages~\cite{prenner2023runbugrun}. 
We conduct experiments to test the impact of the composition of the data and different mixing ratios on  F* code.
\subsection {Synthetic Data Generation Setup}
We select code LLMs demonstrating relatively superior capability in generating F* code: Qwen2.5-Coder-32B-Instruct, Qwen2.5-Coder-14B-Instruct~\cite{hui2024qwen2}, CodeLlama-13b-Instruct-hf~\cite{roziere2023code}, 
DeepSeek-Coder-V2-Lite-Instruct~\cite{liu2024deepseek}, 
DeepSeek-R1-Distill-Qwen-32B ~\cite{guo2025deepseek}. Different prompt templates are adopted in generating different tasks of data. All promptings are done in zero shot. We use a temperature of 0.7 to generate data.  


\paragraph{Evaluation Dataset and Metrics} 
\label{synth simple}
We sample 2,000 instructions from synthesized F* dataset as hold-out test set, equally covering all 3 problem types described above. All evaluations are done on the same test set.
During evaluation, the model first generates an initial response. If incorrect, we prompt the same model again with the erroneous code and the error message for repair.  We record both the initial code generation accuracy and the overall accuracy after a single repair attempt. The response codes are generated with T=0.1. 

\subsection{Results}\label{sec:simplesynthreault}





\begin{table}[]
\centering
\footnotesize
\setlength{\tabcolsep}{1.2pt} 
\setlength{\extrarowheight}{1.2pt}
\resizebox{\columnwidth}{!}{
\begin{tabular}{@{}lcc}
\toprule
\textbf{Model  }                         & \textbf{Pass@1}& \textbf{+Repair} \\ \midrule
\textsc{Qwen-2.5-coder-7B-instruct}      & 0.25             & 0.30                       \\
\textsc{Qwen-2.5-coder-14B-instruct}     & 0.50              & 0.55                        \\
\textsc{Qwen-2.5-coder-32B-instruct}     & 0.48             & 0.58                       \\ 
\textsc{Qwen-2-72B-instruct}            & 0.34             & 0.43                     \\
\textsc{DeepSeek-Coder-33B-Instruct} & 0.29             & 0.38                      \\
\textsc{DeepSeek-Coder-V2-Lite-Instruct} & 0.43             & 0.53                       \\
\textsc{DeepSeek-V3} & 0.66             & 0.78                     \\
\textsc{LLama-3.1-70B }                  & 0.21             & 0.27                         \\
\textsc{Llama-3.3-70B-Instruct}                  & 0.17             & 0.26                     \\
\textsc{GPT-4o}                       &  0.60               &  0.70                       \\        \midrule
\multicolumn{3}{l}{\textbf{Fine-tune Data Mixture}} \\ \midrule
54K F* Only & 0.42 & 0.47 \\
+ Evol & 0.52 &0.56\\
93K F* Only & 0.48 & 0.52\\
+ DSP-V1  & 0.52 & 0.54\\
+ DSP-V1 + Evol + CodeAlpaca + RBR  & 0.58 & 0.62\\
+ DSP-V1 + Evol + CodeAlpaca + RBR (14B)$^{\dagger}$ & \textbf{0.74} & \textbf{0.77}\\
\textit{\space - F* NL2Code} & 0.48 (\textbf{-}) & 0.52(\textbf{-})\\

\bottomrule
\end{tabular}
}
\caption{Performance comparison across different models and fine-tuning data mixtures. \textbf{F* only}: synthetic F* data, Evol: 80K (54K F*) / 50K (93K F*) Magicoder-Evol-Instruct data, \textbf{DSP-V1}: 20K Deepseek-Prover-V1 data, \textbf{CodeAlpaca}: 15K CodeAlpaca data, \textbf{RBR}: 15K RunBugRun data.\\
$^{\dagger}$: Adopting Qwen2.5-Coder-14B as base model.}
\label{tab:syntheticresults}
\end{table}

\paragraph{Comparing Against Powerful LLMs} Compared with 6 popular open-source LLMs: Qwen-2.5-coder-7B-instruct, Qwen-2.5-coder-14B-instruct, Qwen-2.5-coder-32B-instruct ~\cite{hui2024qwen25codertechnicalreport}, Qwen-2-72B-instruct ~\cite{hui2024qwen2}, DeepSeek-Coder-V2-Lite-Instruct~\cite{liu2024deepseek}, LLaMa-3.1-70B, our model achieves the best performance in both generation and repair within the F* framework. The results in Table \ref{tab:syntheticresults} demonstrate that in the initial generation, our model significantly outperforms non-code models such as LLaMA-3.1-70B and Qwen-2-72B in terms of accuracy. At the same time, the accuracy of the generation also surpasses that of code models with larger parameter sizes such as Qwen-2.5-coder-32B-instruct, indicating that our instruction-tuning dataset is highly effective in enhancing the model’s ability to generate F* data. Our initial generation accuracy is also comparable to GPT-4o, which is generally challenging given the size of its base model parameters.
\paragraph{Benefits of data diversity and the effect of different data mixtures }
As shown in Table \ref{tab:syntheticresults}, data diversity has a positive impact on the model's performance. When more diverse language data (e.g. data from Evol-Instruct \& Deepseek-Prover-V1) is added to the F* synthetic dataset, the model’s accuracy on the F* validation set significantly improves, regardless of the amount of original F* synthetic dat (2) Within a certain range, more diverse data leads to better model performance.  (3)  More high-quality synthetic data indeed leads to better model performance, which suggests that for languages like F*, where the model’s knowledge is still limited, increasing the amount of high-quality language-specific fine-tuning data is beneficial for improving the model’s performance. 


\section{Project-Level Proof Synthesis}
In this section, we present the experiments on project-level proof synthesis tasks.
\subsection{Training}

\paragraph{Training Dataset}     
Both the definition generation and repair datasets are formatted using predefined prompt templates, which are listed in Appendix \ref{sec:appendix}. The prompt settings for definition generation and repair are listed in \ref{sec:prompt construction setting}.

Next we integrated the formatted existing and repository-level definition generation data, model-generated and synthetic proof repair data, and the mixed synthetic Function-Level data, and prepared different data mixtures to dine the model, allowing us to explore the influence of each dataset and develop a best training corpus. 

We detail the training set-up for \name in \ref{sec:Model-generate Data Experiment Setting Experiment settings}. 

\subsection{Validation}

After expanding the problem-solution pairs for both definition proof and repair tasks, we apply execution-based validation (detail in \ref{Execution-based data evaluation and dataset filtering}) to filter out incorrect solutions and store error messages for the repair dataset and enhance the diversity of the data. Since our data generation process involves multiple LLMs and stages, and diversity is limited by the fixed number of seed contexts and reference examples, we apply an additional de-duplication step. Specifically, we limit each unique type declaration to at most two instances in the repair dataset, ensuring diversity while minimizing redundancy.

\subsection{\name Auto-pilots and Co-pilots}

In this section, we evaluate the supervised fine-tuned models using different training data mixtures as well as baseline models on the task of proof generation and proof repair. Our evaluation data is a random-sampled, held-out test set with 1K repository -level definition proof problems. The validation process consists of the following three stages:
\begin{table*}[]

\setlength{\tabcolsep}{3pt} 
\centering
\small
\begin{tabular}{@{}l|c c c c@{}}
\toprule
\textbf{Baseline Models}                                     & \textbf{Generate@5} & \textbf{Repair@5} & \textbf{Gen+Rep (Total 10)}& \textbf{Generate@10} \\ \midrule
\textsc{Qwen2.5-Coder-7B-Instruct}                     & 23.4  & 0.2   & 23.6    & 26.3   \\
\textsc{Qwen2.5-Coder-14B-Instruct}                     & 24  & 0.4   & 24.4    & 26.9   \\
\textsc{Qwen2.5-Coder-32B-Instruct}                     & 24.2  & 2.5   & 26.7    & 27.1   \\
\textsc{Deepseek-Coder-V2-LITE-Instruct}                    & 24.4  & 0.7   & 25.1    & 25.1   \\
\textsc{Deepseek-V3}                    & 18.7  & 3.6   & 22.3    & 28.6   \\
\textsc{Deepseek-Coder-33B-Instruct}                    & 22.3  & 4.6   & 26.9    & 28.8   \\
\textsc{GPT-4o}                                       & 22.2  & 1.7   & 23.9    &    23.8    \\
\textsc{Qwen2.5-72B-Instruct}                        & 23.4  & 3.0   & 26.4    & 25.8   \\
\textsc{Llama-3.3-70B-Instruct-Turbo}                  &  19.6     &   3.9    &    23.5     &      21.6 \\
\textsc{Llama-3.1-70B}                  &  19.3     &   2.3    &    21.6     &      22.4\\ \midrule
\multicolumn{5}{l}{\textbf{Data Mixture}} \\ \midrule
\textit{Existing Repos}                                  & 30.7  & 1.0   & 31.7    & 35.3   \\
\hspace{3mm}+ Syn. Project Proof                                   & 32.2  & 2.2   & 34.4    & 36.2   \\
\hspace{3mm} + Func + Syn. Project Proof                          & 32.8  & 2.7   & 35.5    & 37.8   \\
\hspace{3mm}+ Syn. Project Proof  + Syn. Repair                          & 32.7  & 0.7   & 33.4    & 37.5   \\
\hspace{3mm}+ Syn. Project Proof  + Model Repair                          & 33.1  & 4.2   & 37.3    & 37.2   \\
\hspace{3mm}+ Syn. Project Proof  + All Repair                            & \textbf{34.0}  & 4.7   & 38.7    & 38.0   \\ 
\hspace{3mm} \textsc{\name}          & \cellcolor{cyan!10}33.0  & \cellcolor{cyan!10}\textbf{6.4}   & \cellcolor{cyan!10}\textbf{39.4}    & \cellcolor{cyan!10}\textbf{38.5}   \\ 
\bottomrule
\end{tabular}
\caption{Performance comparison of baseline models and fine-tuning data configurations. \textbf{Existing Repos}: 30K existing repository level definition + proofs from the seed dataset; \textbf{Syn. Project Proof} : 30K model generated new definitions + proofs as described in \ref{Generating New Problems}; \textbf{Func}: synthetic simple questions mixed with other datasets in \ref{synth simple}; \textbf{Syn. Repair}: 30K synthetic repair data in \ref{Synthetic Repair Data}, \textbf{Model Repair}: 30K model-generarted repair data in \ref{Repair Data From The Model}; \textbf{All Repair}: Syn. Repair + Model Repair; \textbf{\name}: Existing Repos + Syn. Project Proof + All Repair + 180K mixed function-level coding data used to finetune the best performance in Table \ref{tab:syntheticresults}}
\label{tab:results}
\end{table*}

\begin{figure}
    \centering
    \includegraphics[width=\linewidth]{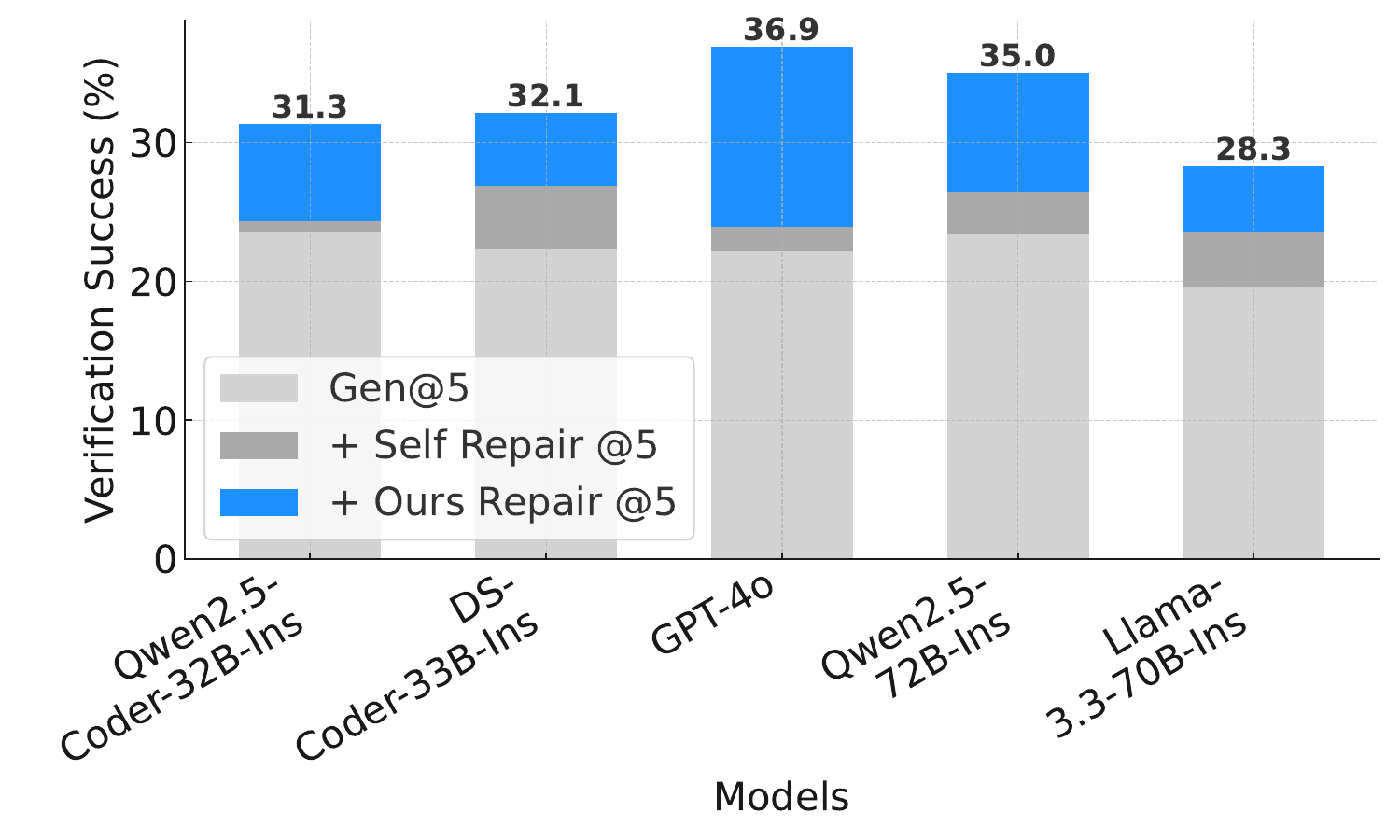}
    \caption{\name repairing failed outputs for state-of-the-art models.}
    \label{fig:repair_result}
\end{figure}


\paragraph{Definition Generation and Self-Repair:}\label{Definition Generation and Self-Repair:} The model generates F* proofs given a type declaration and context (see example prompt in \ref{sec:appendix}), sampling either 5 times followed by 5 repair attempts (Sample-5 + Repair-5) or 10 times directly (Sample-10). If at least one proof compiles, the problem is solved. Otherwise, for the Sample-5 group, failed attempts and error messages are stored for repair, where the model randomly selects an incorrect response and generates 5 additional repair attempts to fix the incorrect proofs. This setup allows us to compare whether self-repairing incorrect generations improves success rates over simply increasing the sampling budget.

\paragraph{Repair Using Outputs from Other Models:} To further assess model performance, we use incorrect proofs generated by a baseline model as input to our fine-tuned model. This allows us to evaluate whether fine-tuning enhances the model’s ability to generalize and improve proof repair across different model outputs.

\subsection{Result}

We evaluate fine-tuning performance using \textbf{Generate@5} and \textbf{Repair@5} metrics. Generate@5 represents the correctness rate when sampling the model five times on the proof generation task, while Repair@5 measures the number of additional questions correctly repaired after sampling five repair attempts on incorrect solutions from the previous stage. As shown in Table \ref{tab:results}, fine-tuning \textsc{Qwen2.5-Coder-14B} on our generated dataset outperforms five larger state-of-the-art coding LLMs in both repository-level proof generation and self-repair tasks.

\paragraph{Proof Repairing Ability:} We use the model fine-tuned with the data mixture Existing Repository-Level Definitions + Synthetic Project Proofs + All Repair dataset (see Table \ref{tab:results}) to repair incorrect outputs from the baseline models. The results in Figure \ref{fig:repair_result} demonstrate its effectiveness in repairing out-of-distribution incorrect proofs. Notably, our fine-tuned model surpasses all baseline large models in correcting their own incorrect answers, bringing a significant improvement in the total number of correct solutions. This suggests a possible application where a fine-tuned, smaller model serves as a debugging assistant for larger models, allowing for potentially efficient co-serving of these powerful models to efficiently adapt to the domain of proof oriented programming.

\paragraph{Data Mixture Analysis:} Table \ref{tab:results} shows how different fine-tuning data mixtures impact proof generation and repair. (1) Augmenting the existing repository-level set with model-generated definitions significantly improves Generate@5 and Repair@5 accuracy, demonstrating the effectiveness of our generated definitions. (2) Rule-based synthetic repair data alone does not enhance repair performance, suggesting that rule-based errors may not accurately reflect the error situations the model would have encountered. (3) Model-generated repair data improves both repair accuracy (from 32.2 to 33.1) and generation accuracy (from 2.7 to 4.2), indicating that language-model-produced errors and repairs can better capture real-world failure patterns than the synthetic ones, that are largely synthetic, generated by mutating ASTs of the program. (4) \name trained with \textbf{Existing Repos + Syn. Project Proof + All Repair + Mixed Function-Level Data} -- our most diverse mixture of data points -- gives the highest Gen+Rep (39.4) and Generate@10 (38.5) across the board, confirming that diverse data improves both proof generation and repair. The function-level synthetic data for F* and mixture from other languages well complement the repository-level F* verification and repair data, and boosts the model's performance.

\section{Related Work}

\paragraph{Language Models For Code}
Large language models (LLMs) have advanced in code generation \cite{chen2021evaluating,austin2021program}, program repair \cite{xia2022less,xia2023automated,jin2023inferfix}, and software engineering tasks like issue fixing \cite{jimenez2023swe} and testing \cite{deng2023large}. Open-source models (e.g., Qwen2.5-Coder \cite{hui2024qwen2}, Deepseek-Coder \cite{guo2024deepseek}) and closed-source models (e.g., GPT-4o \cite{hurst2024gpt}) undergo \textbf{pre-training} on large-scale code datasets \cite{radford2018improving,nijkamp2022codegen}, followed by \textbf{post-training} via instruction fine-tuning \cite{muennighoff2023octopack,roziere2023code,luo2023wizardcoder,codealpaca} or reinforcement learning \cite{ouyang2022training,bai2022training}. While these models excel in common languages like Python and C++, proof-oriented languages such as F* \cite{swamy2011secure} remain underrepresented, limiting their effectiveness in proof synthesis.


\paragraph{Language Models For Formal Proof}
Formal theorem proving and proof repair offer an appealing domain for unlocking the reasoning potential of LLMs, with proofs being easier to verify rigorously without hallucination\cite{yang2023leandojo}, in both mathematical theorem proving and formal program verification. Currently, Language models have shown capability in formal languages such as Isabelle\cite{jiang2022thor,wang2023lego,zhao2023decomposing} and Lean \cite{polu2022formal,han2021proof,yang2023leandojo}. Researchers have also explored various approaches to optimize automated theorem proving using large language models: employing retrieval-augmented assistance\cite{yang2023leandojo}, improving search efficiency by dynamically allocating computational resources \cite{wang2023dt}, predicting the progress of proofs\cite{huang2025leanprogress}, employing LLMs as copilots that assist humans in proving theorems\cite{song2024towards, kozyrev2024coqpilot}, and introducing synthetic data during training\cite{wang2020learning, xin2024deepseek, lin2025goedel, wu2024internlm2}. However, most of these efforts focus on mathematical domains rather than repository-level software verification, which is addressed by PoPilot.

\paragraph{Synthetic Data for Instruction Tuning}
Instruction fine-tuning improves LLMs' ability to follow instructions and relies on high-quality datasets \cite{zhou2024lima,wang2022self}. Since human-annotated datasets are costly \cite{instructgpt2022,kopf2024openassistant,zheng2024lmsyschat1mlargescalerealworldllm}, recent methods focus on LLM-generated instruction data \cite{wang2022self,gunasekar2023textbooks,wang2024codeclm,xu2024magpie}. Self-Instruct \cite{wang2022self} pioneered this approach, later extended by Alpaca \cite{alpaca} and Code Alpaca \cite{codealpaca}. Evol-Instruct \cite{xu2023wizardlm,ahn2024large} and Code Evol-Instruct \cite{luo2023wizardcoder} introduced multi-stage generation for better instruction diversity, though risks of reinforcing biases remain \cite{yu2024large}. OSS-INSTRUCT \cite{wei2024magicoder} and SelfCodeAlign \cite{wei2024selfcodealign} mitigate this by leveraging open-source data, while MultiPL-T \cite{cassano2024knowledge} enables cross-lingual instruction transfer.

\section{Conclusion}
In this work, we propose a synthetic data recipe for instruction-tuning code language models to become proficient proof-oriented programmers in F* under extreme data scarcity. By synthesizing function-level F*, diversifying with other programming languages and tasks and generating new verification tasks on a repository level, we build a powerful \name that outperforms powerful language models, even GPT-4o with only 14B parameter. We further show that \name can work together with existing code LMs to improve their proof-oriented programming capabilities by large margins.
More broadly, we present a variable path for low-resource programming languages and verification tools, lowering the barrier to adopting formal verification in real-world software development.

\paragraph{Limitations}
This work focuses on the programming language of F*, but did not experiment with other languages due to their lack of well-established evaluation suite. 

\bibliography{custom}

\appendix
\section{Experiment Setting}
\subsection{Function-level Data Experiment Setting}\label{sec:Function-level Data Experiment Setting Experiment settings}

We finetune Qwen2.5-Coder-7B on our dataset for one epoch using one NVIDIA A100-40GB GPU. The initial learning rate is set at 5e-6 while the batch size is set at 256. During finetuning, we adopt the OpenRLHF~\cite{hu2024openrlhf} library and modules, applying LoRA~\cite{hu2021lora} with a rank of 32 and an alpha of 32.

\subsection{Prompt Setting}\label{sec:prompt construction setting}

For \textbf{definition generation prompt}, we will provide the type declaration, set the max prompt length to 4096 tokens, and use all the opened modules and premises, 15 selected premises the model may use, and 10 retrieved related examples. For the \textbf{repair prompt}. the problem also includes the incorrect solution and its corresponding error message to guide the model in learning error correction, but excludes the selected premises since \cite{chakraborty2024towards} discovers that the premises has limited effect on proof repairing. We also use fewer related examples in repair prompts to limit the prompt length

\subsection{Model-generate Data Experiment Setting}\label{sec:Model-generate Data Experiment Setting Experiment settings}

We train Qwen2.5-Coder-14B
using supervised fine-tuning for one epoch using 4507
× NVIDIA A100-40GB GPU, with learning rate508
1e-5 and batch size 64. We adopt OpenRLHF (Hu509
et al., 2024), applying LoRA (Hu et al., 2021) with510
a rank of 32 and an alpha of 32.

\section{Problem-Solution Pair Examples}\label{sec:Problem-Solution Pair Examples}

\paragraph{Example 1}

\begin{verbatim}
val clientCertTypeExtension_serializer: 
    LP.serializer 
    c   lientCertTypeExtension_parse

let clientCertTypeExtension_serializer =
    LP.serialize_vlarray 
    1 255 certificateType_serializer 
    1 255 ()
    
\end{verbatim}
\paragraph{Example 2}

\begin{verbatim}
val 
clens_uncompressedPointRepresentation32_x:
LL.clens 
uncompressedPointRepresentation32
  uncompressedPointRepresentation32_X

let clens_uncompressedPointRepresentation32
_x 
: LL.clens 
uncompressedPointRepresentation32 
uncompressedPointRepresentation32_X = {
  LL.clens_cond = (fun _ -> True);
  LL.clens_get = (fun x -> x.x);
}
     
    
\end{verbatim}

\section{Prompt Templates}
\label{sec:appendix}

\subsection{Definition Generation Prompt Example}
\label{Definition Generation Prompt Example}

\begin{verbatim}
You are tasked with F* code generation. 
You will be given a type declaration, 
and you need to write a definition for 
it.
## Type declaration:
val tau: Prims.unit -> Tac unit

	1. Write the definition that satisfies the 
    above type. 
	2. Start the definition with ``` let tau ``` .
	3. Only write in F* code.
	4. Add END token after completing the 
    definition.
    
## Possibly useful premises:

	FStar.Tactics.Effect.raise
	FStar.Pervasives.reveal_opaque
	FStar.Tactics.Effect.get
	FStar.Tactics.Effect.tactic
	FStar.Pervasives.Native.snd
	FStar.Pervasives.Native.fst
	FStar.Monotonic.Pure.
    elim_pure_wp_monotonicity
	FStar.Tactics.Types.issues
	FStar.Pervasives.dfst
	FStar.Pervasives.dsnd
	FStar.Tactics.Effect.tac_return
	FStar.Monotonic.Pure.
    elim_pure_wp_monotonicity_forall
	FStar.Tactics.Effect.tac
	FStar.Monotonic.Pure.
    intro_pure_wp_monotonicity
	Prims.l_True
    
## Already opened files and delared modules

	open FStar
	open FStar.Pervasives
	open Prims
	open FStar.Tactics.V2
    
## Related types and definitions

val tau: Prims.unit -> Tac unit
let tau () : Tac unit =
    apply_lemma (`refl)
val tau: Prims.unit -> Tac unit
let tau () : Tac unit =
    let * = implies*intro () in
    let * = implies*intro () in
    let * = implies*intro () in
    let b = implies_intro () in
    var_retype b; // call retype, 
    get a goal `p == ?u`
    let pp = `p in
    let rr = `r in
    grewrite pp rr; // rewrite p to q, 
    get `q == ?u`
    trefl (); // unify
    apply_lemma (`l); //prove (p == q), 
    asked by grewrite
    let e = cur_env () in
    match vars_of_env e with
    | [_;_;_;b] ->
        let t = type_of_binding b in
        let t = norm_term [] t in 
        // contains uvar redexes.
        if FStar.Order.ne 
        (compare_term t rr)
        then fail "binder was not retyped?"
        else ();
        apply_lemma (`l2);
        assumption' ();
        qed ()
    | _ ->
        fail "should be impossible"


Write your response below.


\end{verbatim}

\subsection{Repair Prompt Example}

\begin{verbatim}
You are tasked with F* code generation. 
You will be given a type declaration, 
and an incorrect student solution. 
You need to produce a correct solution.

## Type declaration:

val clientHelloExtension_e_session
_ticket_clens:
    LL.clens clientHelloExtension_
    e_session_ticket
  sessionTicket

	1. Write the definition 
    that satisfies the above type. 
	2. Start the definition 
    with 
    ``` 
let clientHelloExtension
_e_session_ticket_clens 
    ``` .
	3. Only write in F* code.
	4. Add END token after completing 
    the definition.



## Already opened files and delared modules

	open MiTLS.Parsers.SessionTicket
	open Prims
	open FStar.Bytes
	open MiTLS.Parsers
	open FStar.Pervasives
	open FStar


## Related types and definitions

val newSessionTicketExtension_clens'
_session_ticket:
LL.clens newSessionTicketExtension
  newSessionTicketExtension_e_default
let 
newSessionTicketExtension_clens'
_session_ticket 
: LL.clens newSessionTicketExtension 
newSessionTicketExtension_e_default = 
LL.clens_dsum_payload 
newSessionTicketExtension_sum 
(LL.Known 
(known_extensionType_as_enum_key 
Session_ticket))

val newSessionTicketExtension_clens'
_client_certificate_type:
LL.clens newSessionTicketExtension
  newSessionTicketExtension_e_default
let newSessionTicketExtension_clens'
_client_certificate_type : 
LL.clens newSessionTicketExtension 
newSessionTicketExtension_e_default 
= LL.clens_dsum_payload 
newSessionTicketExtension_sum 
(LL.Known (known_extensionType_as_enum_key
Client_certificate_type))



## Student Solution

@@ Student F* Code
```fstar
open FStar
open Prims
open FStar.Pervasives
open MiTLS.Parsers
open MiTLS.Parsers
open FStar.Bytes
module U8=FStar.UInt8
module U16=FStar.UInt16
module U32=FStar.UInt32
module U64=FStar.UInt64
module LP=LowParse.Spec.Base
module LS=LowParse.SLow.Base
module LSZ=LowParse.SLow.Base
module LPI=LowParse.Spec.AllIntegers
module LL=LowParse.Low.Base
module L=FStar.List.Tot
module B=LowStar.Buffer
module BY=FStar.Bytes
module HS=FStar.HyperStack
module HST=FStar.HyperStack.ST
open MiTLS.Parsers.SessionTicket
open MiTLS.Parsers.
ClientHelloExtension_e_session_ticket
#push-options "--initial_fuel 2 -
-max_fuel 8 --initial_ifuel 1 
--max_ifuel 2 --smtencoding.elim_box false 
--smtencoding.nl_arith_repr boxwrap 
--smtencoding.l_arith_repr boxwrap 
--smtencoding.valid_intro true 
--smtencoding.valid_elim false 
--z3rlimit 5 --z3rlimit_factor 
1 --z3seed 0"

#restart-solver
val 
clientHelloExtension_e_session_ticket_clens
:LL.clens 
clientHelloExtension_e_session_ticket
// Error Range Start - Line 27
  sessionTicket 
// Error Range End - Line 27
let 
clientHelloExtension_e_session_ticket_clens
:LL.clens sessionTicket =
  {
    LL.clens_cond = (fun _ -> True);
    LL.clens_get
    =
    (fun (x: 
    clientHelloExtension_e_session_ticket) 
    -> (x <: sessionTicket))
  }

@@ Error Message
  - Expected type "Type"; but 
  "LL.clens sessionTicket"
  has type "t2: Type -> Type"


  - Expected type "Type"; 
  but "LL.clens sessionTicket"
  has type "t2: Type0 -> Type"


  - Expected type 
  "LL.clens sessionTicket"; 
  but "LL.Mkclens (fun _ -> l_True) 
  (fun x -> x <: sessionTicket)" has type 
  "LL.clens 
  clientHelloExtension_e_session_ticket 
  sessionTicket"


  - Expected type 
  "LL.clens 
  clientHelloExtension_e_session_ticket 
  sessionTicket"; 
  but "LL.Mkclens (fun _ -> l_True) 
  (fun x -> x <: sessionTicket) 
  <: LL.clens sessionTicket" 
  has type "LL.clens 
  sessionTicket"


Write your response below.

\end{verbatim}

\subsection{New Definition Prompt Example}\label{sec:New Definition Prompt Example}

\begin{verbatim}
You are tasked with generating F* code. 
You will be given some premises, 
opened modules and some example type 
declarations and definitions. Your 
goal is to write a different type 
declaration and a corresponding 
definition that satisfies the 
type declaration.

You can use the information 
provided in the following sections 
to help you construct the new 
type declaration and definition. 
Here's how you can use each section:

Possibly useful premises: 

This section lists modules, types, 
or functions that might be helpful. 
Consider incorporating them into 
your definition if appropriate.

Already opened files and declared 
modules: 

The modules listed here 
are already opened in the context. 
You can use definitions from these
modules directly without needing 
to prefix them with the module name.

Example type declarations and 
definitions: 

This shows some examples 
of how a definition satisfying a 
similar type declaration can be 
written. Each example is delimited 
using "```". Use this as a reference 
for the structure and style, 
but do not use the examples 
definitions in your answer.

Use the information provided to 
write one new type declaration 
and a definition that satisfies 
this new type declaration. 
Only write in F* code, you don't 
need to provide any explanation or 
example.
Start your new type declaration 
and definition with "val" and "let" 
respectively, and add END after 
completing the definition. 
You should only use the premises 
from "Possibly useful premises".



## Possibly useful premises:

FStar.Tactics.Effect.raise
FStar.Pervasives.Native.fst
FStar.Pervasives.Native.snd
FStar.Tactics.Types.issues
FStar.Tactics.Effect.get
FStar.Pervasives.dfst
FStar.Pervasives.dsnd
GradedMonad.monoid_nat_plus
GradedMonad.st
FStar.Pervasives.st_post_h
FStar.Pervasives.reveal_opaque
FStar.Issue.mk_issue
FStar.Pervasives.st_post_h'
FStar.Pervasives.st_pre_h
FStar.Monotonic.
Pure.elim_pure_wp_monotonicity

## Already opened files 
and declared modules

	open FStar.Pervasives
	open FStar
	open Prims



## Example definitions

```
val st_monad (s: _) 
: monad (st s)
instance st_monad s 
: monad (st s) =
{
   return = (fun 
   #a (x:a) -> 
   (fun s -> x, s));
   bind   = 
   (fun #a #b (f: st s a) 
   (g: a -> st s b) (s0:s) ->
           let x, s1 = f s0 in
           g x s1);
}
```

```
val monad_functor (#m: _) 
(d: monad m) : functor m
instance monad_functor #m 
(d : monad m) : functor m =
  { fmap = (fun #_ #_ f x 
  -> bind #m x (fun xx 
  -> return #m (f xx))); }
```

```
val FStar.DM4F.MonadLaws.st 
= s: Type -> a: Type -> Type
let st (s:Type) (a:Type) 
= s -> Tot (a * s)
```

```
[@@ FStar.Tactics.
Typeclasses.tcinstance]
val opt_monad:monad option
instance opt_monad : 
monad option =
{
   return = (fun #a 
   (x:a) -> Some x);
   bind = (fun #a 
   #b (x:option a) 
   (y: a -> option b) ->
         match x with
         | None -> None
         | Some a -> y a)
}
```

Write your response below.
\end{verbatim}
\begin{figure}
    \centering
    \includegraphics[width=1.06\linewidth]{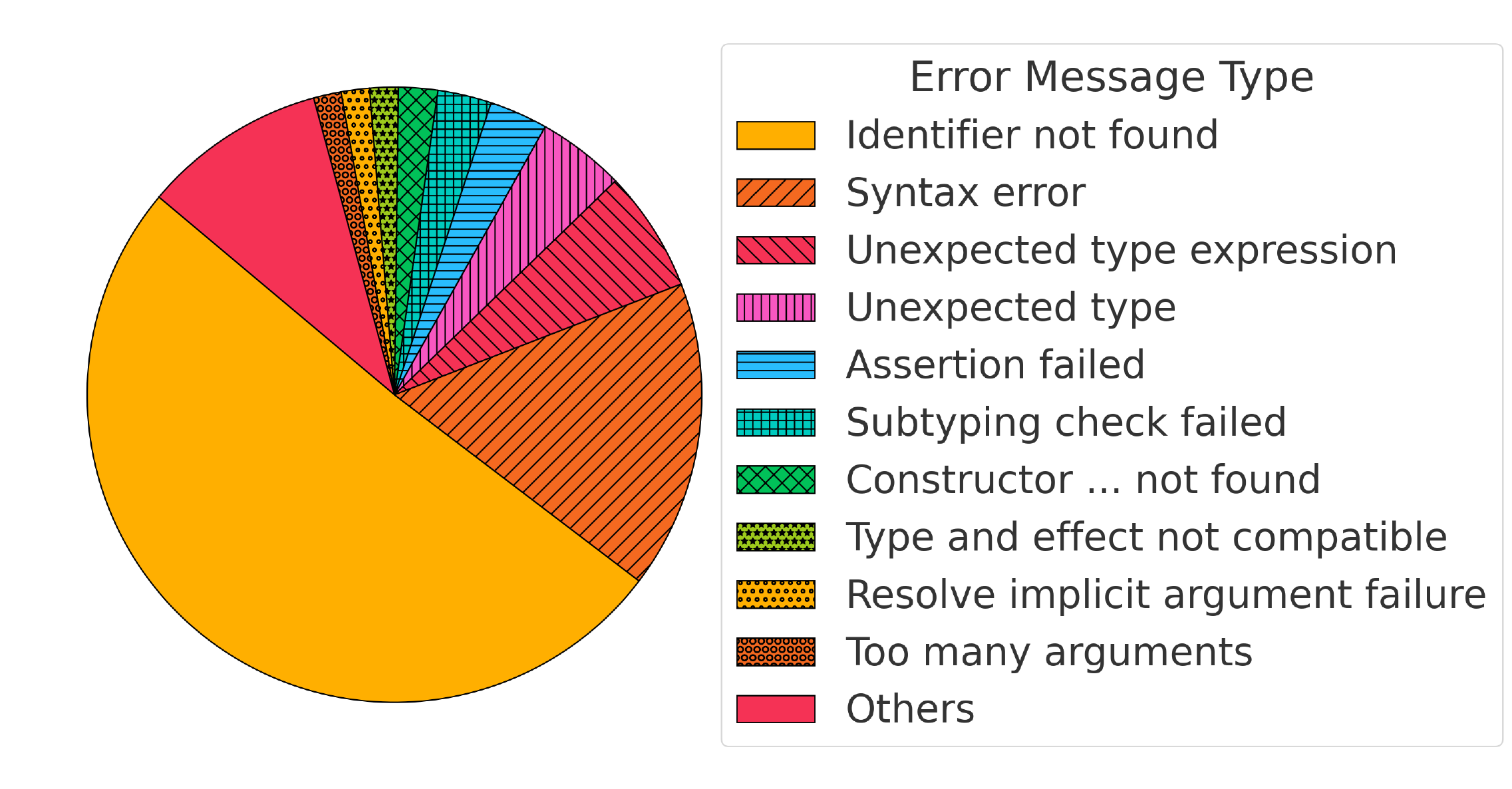}
    \caption{Distribution of Top 10 Error Types of Model-Generated Repair Data.}
    \label{fig:error-distribution}
\end{figure}

\section{Ablation Study}

\subsection{Repair Sampling Strategy Ablation study}

In the self-repair experiment, we tested an alternative strategy: sampling 5 incorrect solutions and attempting to repair each once (totaling 5 repair attempts). However, this approach performed significantly worse than the Sample 1 \& Repair 5 strategy, so we chose the latter in the following experiments. The comparison of both strategies on three models is shown in Table \ref{tab:sampling_strategy}.

We hypothesize that proof repair remains a challenging task, even for our fine-tuned model. Allowing multiple repair attempts on the same problem increases the chances of success, leading to better overall accuracy.
\begin{table}[t]
\small
\setlength{\tabcolsep}{2pt} 
\centering
\begin{tabular}{@{}c|c|c|c@{}}
\toprule
\textbf{Model}               & \textbf{Gen@5} & \textbf{sample1-on-5} &  \textbf{sample5-on-1} \\ 
\midrule
\textsc{Our Best Model}       &  34 & +1.7    &  +4.7 \\
\textsc{Qwen2.5-Coder-32B}       & 23.5  & +0.8   &  +7.8 \\
\textsc{DS-Coder-33B}      & 22.3  & +4.6   &  +9.8 \\
\bottomrule
\end{tabular}
\caption{Comparison of repair sampling strategies: \textbf{sample1-on-5} repairs each incorrect solution once, while \textbf{sample5-on-1} repairs the same incorrect solution multiple times.}
\label{tab:sampling_strategy}
\end{table}

\subsection{Fine-tune on smaller model}

We used the same data of \name to fine-tune a Qwen/Qwen2.5-Coder-7B model, the result is shown in table ~\ref{tab:small} with comparison to \name and the baseline models. 

Given the small size of \name, the performance is reasonably good. However, the repair capability is still minor, so we will just report \name in our main part. 
\begin{table*}[]
\setlength{\tabcolsep}{3pt} 
\centering
\small
\begin{tabular}{@{}l|c c c c@{}}
\toprule
\textbf{Baseline Models}                                     & \textbf{Generate@5} & \textbf{Repair@5} & \textbf{Gen+Rep (Total 10)}& \textbf{Generate@10} \\ \midrule
\textsc{Qwen2.5-Coder-32B-Instruct}                     & 23.5  & 0.8   & 24.3    & 27.1   \\
\textsc{Deepseek-Coder-33B-Instruct}                    & 22.3  & 4.6   & 26.9    & 28.8   \\
\textsc{GPT-4o}                                       & 22.2  & 1.7   & 23.9    &    23.8    \\
\textsc{Qwen2.5-72B-Instruct}                        & 23.4  & 3.0   & 26.4    & 25.8   \\
\textsc{Llama-3.3-70B-Instruct-Turbo}                  &  19.6     &   3.9    &    23.5     &      21.6  \\ \midrule
\multicolumn{5}{l}{\textbf{Data Mixtures (\textit{Qwen/Qwen2.5-Coder-7B})}} \\ \midrule
\textit{Existing Repos}                                  & 12.9  & 3.6   & 16.5    & 18.5   \\
\hspace{3mm}+ Syn. Project Proof                                   & 14.2  & 3   & 17.2    & 19.8   \\
\hspace{3mm} + Func + Syn. Project Proof                          & 14  & 3.7   & 17.7    & 18.7   \\
\hspace{3mm}+ Syn. Project Proof  + Syn. Repair                          & 13.9  & 3.8   & 17.7  & 18.7   \\
\hspace{3mm}+ Syn. Project Proof  + Model Repair                          & 15.2  & 4.4   & 19.6 & 20.8   \\
\hspace{3mm}+ Syn. Project Proof  + All Repair                            & 15  & 4.7   & 19.7  & 21.2   \\
\textsc{\name-small}                                 & 21.9  & 3.9   & 25.8    & 29.2   \\
\hspace{3mm} \textsc{\name}          & \cellcolor{cyan!10}\textbf{33.0}  & \cellcolor{cyan!10}\textbf{6.4}   & \cellcolor{cyan!10}\textbf{39.4}    & \cellcolor{cyan!10}\textbf{38.5}   \\ 
\bottomrule
\end{tabular}
\caption{Performance comparison of the small model}
\label{tab:small}
\end{table*}

 


\end{document}